%% file: main.tex
\documentclass[conference, 9pt]{IEEEtran}

\IEEEoverridecommandlockouts

\input{misc/packages.tex}
\input{misc/tikz.tex}

\begin{document}
\title{Attacks on Robust Distributed Learning Schemes via Sensitivity Curve Maximization}

\author{\IEEEauthorblockN{
Christian A. Schroth\IEEEauthorrefmark{1},
Stefan Vlaski\IEEEauthorrefmark{2} and
Abdelhak M. Zoubir\IEEEauthorrefmark{1}}
\IEEEauthorblockA{\IEEEauthorrefmark{1}Signal Processing Group, Technische Universität Darmstadt, Germany}
\IEEEauthorblockA{\IEEEauthorrefmark{2}Department of Electrical and Electronic Engineering, Imperial College London, UK}
\IEEEauthorblockA{\IEEEauthorrefmark{1}\{schroth, zoubir\}@spg.tu-darmstadt.de,
\IEEEauthorrefmark{2}s.vlaski@imperial.ac.uk}
\thanks{The work of Christian A. Schroth and Abdelhak M. Zoubir has been funded by the LOEWE initiative (Hesse, Germany) within the emergenCITY centre and by DFG Project under Grant 431431951 / ZO 215/19-1.}}

\maketitle
\setstretch{1} % max 83

\begin{abstract}
Distributed learning paradigms, such as federated or decentralized learning, allow a collection of agents to solve global learning and optimization problems through limited local interactions. Most such strategies rely on a mixture of local adaptation and aggregation steps, either among peers or at a central fusion center. Classically, aggregation in distributed learning is based on averaging, which is statistically efficient, but susceptible to attacks by even a small number of malicious agents. This observation has motivated a number of recent works, which develop robust aggregation schemes by employing robust variations of the mean. We present a new attack based on sensitivity curve maximization (SCM), and demonstrate that it is able to disrupt existing robust aggregation schemes by injecting small, but effective perturbations. 
\end{abstract}


\input{sections/intro.tex}

\input{sections/learning.tex}

\input{sections/attack.tex}
\input{sections/simulation.tex}

\input{sections/conclusion.tex}

%\section*{Acknowledgment}

\printbibliography

\end{document}

%% file: misc/packages.tex
\usepackage{mathtools, amssymb}
\usepackage{subdepth} %forces subscripts to be on same height
\usepackage{dsfont}
\usepackage{setspace} % reduce line sapcing
\let\labelindent\relax
\usepackage{enumitem}

\usepackage{tikz}
\usepackage{pgfplots,etoolbox}
\usetikzlibrary{pgfplots.groupplots}
\usetikzlibrary{math}
\pgfplotsset{compat=newest}
\usepgfplotslibrary{external}

\usepackage{subcaption}

\usepackage{float}
\usepackage[ruled,vlined]{algorithm2e}% http://ctan.org/pkg/algorithm2e
\DontPrintSemicolon

\usepackage[backend=biber, style=ieee, isbn=false, doi=false]{biblatex}

\bibliography{refs.bib}
\makeatletter
\newcommand{\removelatexerror}{\let\@latex@error\@gobble}
\makeatother

\DeclareMathOperator*{\argmax}{arg\,max}

%% file: misc/tikz.tex
\newlength{\figurewidth}
\newlength{\figureheight}

\definecolor{matlabblue}{rgb}{0 0.4470 0.741}
\definecolor{matlaborange}{rgb}{0.8500 0.3250 0.0980}
\definecolor{matlabyellow}{rgb}{0.9290 0.6940 0.1250}
\definecolor{matlabpurple}{rgb}{0.4940 0.1840 0.5560}
\definecolor{matlabgreen}{rgb}{0.4660 0.6740 0.1880}
\definecolor{matlablightblue}{rgb}{0.3010 0.7450 0.9330}
\definecolor{matlabred}{rgb}{0.6350 0.0780 0.1840}

\pgfplotscreateplotcyclelist{matlabcolor}{%
	matlabblue, mark=diamond*\\%
	matlaborange, mark=*\\%
	matlabyellow, mark=square*\\%
	matlabpurple, mark=triangle*\\%
	matlabgreen, mark=star\\%
	matlablightblue\\%
	matlabred\\%
}

%% file: sections/intro.tex
\section{Introduction}

Distributed learning paradigms, such as federated or decentralized learning are growing in importance as data is increasingly available in dispersed locations, while central aggregation of data is infeasible due to concerns around privacy or communication efficiency\footnote{We use ``distributed'' for any structure where data remains local at individual agents, which includes federated and decentralized architectures. The term ``decentralized'' is used for a network without fusion center. In the literature, ``distributed'' and ``decentralized'' are sometimes used interchangeably.}. In cooperative settings, well-designed distributed algorithms can match the performance of a centralized benchmark, which has access to all data at a single location~\cite{Chen.2015, Sayed.2014, Lian.2017, Nedic.2018}. At the same time, however, their reliance on linear averaging for model aggregation renders most of these strategies susceptible to attacks by even a small number of deviating agents. This observation has motivated the development of robust aggregation schemes for distributed learning, primarily focusing on federated network structures. Popular methods are based on robust variations of the mean, such as the trimmed mean and median \cite{Yin.2018} or weighted geometric median \cite{Wang.2023}, RFA~\cite{Pillutla.2022b}, Krum~\cite{Blanchard.2017} and Bulyan~\cite{Mhamdi.2018}. A recent survey can be found in~\cite{RodriguezBarroso.2023}.

Compared to federated learning, the literature of robust decentralized learning is more sparse. ByRDiE~\cite{Yang.2019} and BRIDGE~\cite{Fang.2019} use a coordinate-wise trimmed mean, where the latter has variants using a coordinate-wise median, Krum or a combination thereof. The Iterative Outlier Scissor (IOS)~\cite{Wu.2022}, iteratively discards a number of weights, which are furthest away from the weighted average. The number of discarded weights is equal to the number of estimated Byzantine outliers. In~\cite{Vlaski.2022b}, a robust and efficient method for element-wise robust aggregation based on the Biweight Tukey M-estimator was presented. Self Centered Clipping (SCC)~\cite{He.2022} is a variant of the trimmed-mean where the distance is measured from the local nodes' own weight and weights beyond a specified threshold are clipped. All of these aggregation schemes have in common that they rely in one way or another on a distance measure to detect and reject outliers. 

These schemes are quite effective against simple and/or non-intelligent attacks, but a byzantine attacker which is aware of the used aggregation scheme, will be able to exploit this information and will intelligently attack the aggregation scheme. As it was shown in ``A little is enough'' (ALIE)~\cite{Baruch.2019b}, most robust federated learning methods can be disturbed by injecting a carefully selected value, which is small enough to be not detected by the defense mechanism, but large enough to disturb the estimation. \cite{Fang.2019b} proposed a method, where the malicious nodes try to generate weights, which will lead to the largest deviation in the inverse direction. In \cite{Xie.2020}, it is shown that the inner product between the true gradient and the robust estimator has to be non-negative, otherwise a malicious agent could invert the direction of the gradient. More attack mechanisms are presented in \cite{Wang.2020, Lyu.2020}.

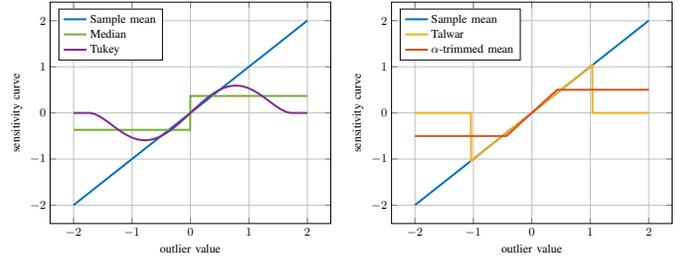
\begin{figure}
	\centering
	\subcaptionbox{SCs of sample mean, median and Biweight Tukey\label{fig:SC1}}[.49\columnwidth]{\resizebox{0.49\columnwidth}{!}{\input{figs/SC1.tex}}}
	\hfill
	\subcaptionbox{SCs of sample mean, Talwar and $\alpha$-trimmed mean \label{fig:SC2}}[.49\columnwidth]{\resizebox{0.49\columnwidth}{!}{\input{figs/SC2.tex}}}
	\caption{Overview of sensitivity curves (SCs) for different aggregation schemes. Tukey and the trimmed means are tuned, such that they achieve 95\% efficiency.}
	\label{fig:SC}
	\vspace{-5mm}
\end{figure}

A novel defense against malicious attacks is MixTailor~\cite{AliRamezaniKebrya.2022}, which does not rely on one static aggregation scheme, but randomly draws an aggregation scheme from a set of predefined schemes. This prevents even omniscient attackers from crafting a perfect outlier, as they only could inject an optimal outlier in expectation.

Motivated by the observations of ALIE, we develop an attack scheme on robust aggregators by maximizing their sensitivity curve (SC). The sensitivity curve describes the influence of an outlier on the estimated value. The two extremes can be observed in Figure~\ref{fig:SC1}, where for the sample mean the influence of an outlier increases linearly with its value, while for the median the value of an outlier has almost no influence. In what follows, we will develop an attack scheme which maximizes the influence on the SC. Our contributions are:
\begin{itemize}
	\item We propose a novel attack scheme for a broad range of robust aggregation schemes, by tailoring perturbations to the sensitivity curve of the aggregator. 
    \item We demonstrate numerically that the proposed scheme is able to disrupt most popular aggregation schemes in the literature, even those based on robust variations of the mean, such as the median.
\end{itemize}

The remainder of the paper is organized as follows. Decentralized learning, different robust aggregation schemes and the sensitivity curve are introduced in Section~\ref{sec:learn}. Our proposed attack scheme is given in Section~\ref{sec:attack}. Section~\ref{sec:sim} gives an overview of the experimental validations and finally, a conclusion and outlook are given in Section~\ref{sec:con}.

%% file: figs/SC1.tex
\begin{tikzpicture}
	\begin{axis}[
		width = \figurewidth,
		height = \figureheight,
		xlabel = outlier value,
		ylabel = sensitivity curve,
		grid=both,
		cycle list name=matlabcolor,
		legend pos=north west,
		legend cell align={left},
		]
		\addplot+[thick,matlabblue,line width=1.5pt,no marks] table[x index = 0,y index=1,col sep=comma] {figs/SC.csv};
		\addplot+[thick,matlabgreen,line width=1.5pt,no marks] table[x index = 0,y index=2,col sep=comma] {figs/SC.csv};
		\addplot+[thick,matlabpurple,line width=1.5pt,no marks] table[x index = 0,y index=3,col sep=comma] {figs/SC.csv};
		\legend{Sample mean, Median, Tukey};
	\end{axis}
\end{tikzpicture}

%% file: figs/SC2.tex
\begin{tikzpicture}
	\begin{axis}[
		width = \figurewidth,
		height = \figureheight,
		xlabel = outlier value,
		ylabel = sensitivity curve,
		grid=both,
		%xmin=87,
		%xmax=105,
		%cycle list name=matlabcolor,
		legend pos=north west,
		legend cell align={left},
		]
		\addplot+[thick,matlabblue,line width=1.5pt,no marks] table[x index = 0,y index=1,col sep=comma] {figs/SC.csv};
		\addplot+[thick,matlabyellow,line width=1.5pt,no marks] table[x index = 0,y index=5,col sep=comma] {figs/SC.csv};
		\addplot+[thick,matlaborange,line width=1.5pt,no marks] table[x index = 0,y index=4,col sep=comma] {figs/SC.csv};
		\legend{Sample mean, Talwar,  $\alpha$-trimmed mean};
	\end{axis}
\end{tikzpicture}

%% file: sections/learning.tex
\section{Preliminaries}
\label{sec:learn}

\subsection{Decentralized Learning}
We will consider a collection of \( K \) agents, and associate with each agent a local objective function
\begin{align}
    J_k(w) = \mathds{E} Q(w; \boldsymbol{x}_{k})
\end{align}
where \( w \) denotes the parameters or weights of a model maintained by each agent, and \( \boldsymbol{x}_k \) is a random variable representing the data at agent \(k \). Most federated and decentralized learning algorithms aim to pursue an optimal model according to the consensus optimization problem
\begin{align}
    J(w) = \frac{1}{K} \sum_{k=1}^K J_k(w).
\end{align}
In decentralized learning, agents only exchange intermediate estimates on a peer-to-peer basis, without communicating with a central fusion center. For example, the ATC-diffusion algorithm takes the form~\cite{Chen.2015, Sayed.2014}
\begin{align}
	\boldsymbol{\phi}_{k,i} =&\: \boldsymbol{w}_{k,i-1} - \mu \widehat{\nabla J}_{k}(\boldsymbol{w}_{k,i-1})\label{eqn:adapt}\\
	\boldsymbol{w}_{k,i} =&\: \sum_{\ell \in \mathcal{N}_{k}} a_{\ell k} \boldsymbol{\phi}_{\ell,i}
	\label{eqn:average}
\end{align} 
where $\mu > 0$ denotes the step-size and $\mathcal{N}_{k}$ denotes the neighborhood of node $k$, including node $k$. The local adaptation~\eqref{eqn:adapt} is driven by a stochastic gradient approximation \( \widehat{\nabla J}_k(\boldsymbol{w}_{k, i-1})\). It is common to choose as \( \widehat{\nabla J}_k(\boldsymbol{w}_{k, i-1}) = \nabla Q(\boldsymbol{w}_{k, i-1}; \boldsymbol{x}_{k, i})\), where \( \boldsymbol{x}_{k, i}\) denotes the sample available to agent \( k \) at time \( i \). Commonly, the aggregation step in~\eqref{eqn:average} is a non-robust sample mean. The choice of the sample mean is well justified in benign scenarios without malicious agents, as it usually exhibits the highest statistical efficiency, and hence the fastest rate of convergence. In scenarios with malicious agents, this is susceptible to significant degeneration of the performance. By replacing the non-robust average in Equation~\eqref{eqn:average} with a general aggregation rule \( \mathrm{AGG}\left( \cdot \right) \) as in
 \begin{align}
	\boldsymbol{w}_{k,i} = \text{AGG}(\boldsymbol{\phi}_{\ell,i}), \quad \ell \in \mathcal{N}_{k},
	\label{eqn:robAGG}
\end{align} 
it is possible to introduce a variety of different aggregation schemes, including schemes which are robust against outliers. 

\subsection{Aggregation Schemes}
For simplicity we will limit the investigation in this paper to element-wise aggregation schemes. For a set of samples $\mathcal{Y} = \{y_{1},  y_{2}, \dots, y_{N}\}$:
\begin{enumerate}[leftmargin=*]
	\item \textit{Sample mean:}
	\begin{equation}
		\hat{\mu} = \frac{1}{|\mathcal{Y}|}\sum_{y_n \in \mathcal{Y}} y_{n}
	\end{equation}
	Achieves a breakdown point of \( 0 \), meaning that a single malicious sample can result in arbitrary deterioration of performance~\cite{Zoubir.2018}. This choice of aggregator recovers~\eqref{eqn:average}.
	\item \textit{Median:} Select the median of the samples in \(\mathcal{Y}\). Achieves a breakdown point of \(0.5\), meaning that \( 50\%\) of samples can be corrupted before the median breaks down. This choice of aggregator recovers~\cite{Yin.2018}.
	\item \textit{$\alpha$-trimmed mean:} The $\alpha N$ largest and smallest values are discarded, resulting in a total $2\alpha  N$ values being removed. Achieves a breakdown point of $\alpha$. This choice of aggregator is employed in~\cite{Yin.2018, Yang.2019, Wu.2022}.
	\item \textit{Talwar \cite{Menezes.2021}:} Also called Huber type-skipped mean \cite{Hinich.1975}, falls into the class of M-estimators which solve for a location estimate
	\begin{equation}
		\sum_{y_n \in \mathcal{Y}} \psi\left(\frac{y_{n} - \hat{\mu}}{\hat{\sigma}}\right) = 0
	\end{equation}
	iteratively with a fixed point algorithm for $\hat{\mu}$ starting with a robust initial location and scale estimate and with 
	\begin{equation}
		\psi(x)= 
		\begin{cases}
			x & ,|x|\leq c\\
			0 & ,|x|> c.
		\end{cases}
		\label{eqn:psiTal}
	\end{equation}
    As initial estimates commonly the median absolute deviation (mad) and median are chosen.
	\item \textit{Biweight Tukey:} The well known M-estimator \cite[p.~11]{Zoubir.2018} with
	\begin{equation}
		\psi(x)= 
		\begin{cases}
			x\left(1-\frac{x^2}{c^2}\right)^{2} &,|x|\leq c\\
			0 &,|x|> c.
		\end{cases}
		\label{eqn:psiTuk}
	\end{equation}
    achieves a breakdown point of close to 0.5, while being more efficient than the median. This type of aggregator is considered in~\cite{Vlaski.2022b}.
\end{enumerate}

\subsection{Sensitivity Curve}
Each aggregator can be associated with a sensitivity curve (SC), which measures the influence a single maliciously designed sample can exhibit on the estimator. We will use these sensitivity curves to determine worst case expressions for malicious perturbations. Formally, the SC or empirical influence function~\cite{Hampel.1986} for an estimator $\text{AGG}(\cdot)$ and samples $\mathcal{Y} = \{y_{1},  y_{2}, \dots, y_{N-1}\}$ of length $N-1$ is defined as
\begin{align}
	\text{SC}(\mathcal{Y}, z) = N (\text{AGG}(\mathcal{Y} \cup z) - \text{AGG}(\mathcal{Y})).
\end{align}
It describes the bias of the estimator when an additional observation $z$ is added to a sample of size $N-1$. Examples of SCs of the considered aggregation schemes are shown in Figure~\ref{fig:SC}.

For our further investigations, it is useful to extend the definition of the SC to account for $P$ identical outliers $\mathcal{Z} = \{z \cdot \mathds{1}_{P}\}$ where $\mathds{1}_{P}$ is a repeater function and with $\mathcal{Y} = \{y_{1},  y_{2}, \dots, y_{N-P}\}$ as
\begin{align}
	\text{SC}(\mathcal{Y}, \mathcal{Z}) = N (\text{AGG}(\mathcal{Y} \cup \mathcal{Z})- \text{AGG}(\mathcal{Y})).
\end{align}
We will use this definition to quantify the effect that \( P \) malicious agents can have when coordinating to induce maximum bias in the aggregation. We note that the SC is not necessarily fixed, but is a function on the underlying random samples, e.g. for the median, the SC depends on the distance between the random samples and therefore can take larger or smaller values. But the fundamental shape is determined by the aggregator and is not influenced by the malicious samples.

%% file: sections/attack.tex
\section{Proposed Sensitivity Curve Maximization Attack}
\label{sec:attack}

In \cite{Baruch.2019b} the ALIE attack model for federated learning is proposed, which illustrates the fact that most robust aggregation rules detect and reject outliers by a distance measure. 
Assuming that the samples, sent to node $k$ from its neighborhood, are normally distributed, ALIE estimates the mean and variance of this distribution. Equipped with those estimates the cdf is used to estimate a value, which will lead to the rejection of benign nodes. The number of rejected benign nodes is chosen, such that the malicious nodes will gain the majority in the neighborhood.

Following the idea of injecting small values, which will ultimately lead to an estimation bias and hence, a breakdown, we propose an attack model which is based on sensitivity curve maximization (SCM). As introduced in the previous section, the SC describes the influence of an outlier on the estimate. Finding the minimal value, which maximizes the SC and injecting this value into the aggregation, will lead to the greatest possible distortion of the estimate. To estimate this optimal value $z_{m}^{\textrm{opt}}$ an omniscient byzantine attacker is assumed, but it should be possible to only use the estimates of the malicious nodes, to gain an estimated optimal attack value.

The set of all benign values for agent $k$ in dimension $m$ is defined as $\mathcal{Y}_{m} = \{\boldsymbol{\phi}_{l,i}(m)\}_{l \in \mathcal{N}^{b}_{k}}$ and the set of malicious values is defined as $\mathcal{Z}_{m} = \{z_{m} \cdot \mathds{1}_{|\mathcal{N}^{m}_{k}|}\}$, where $\mathcal{N}^{b}_{k}$ and $\mathcal{N}^{m}_{k}$ denote the benign neighborhood and the malicious neighborhood, respectively. Finding the smallest value $z^{\textrm{opt}}_{m}$, which maximizes the SC, requires solving
%\begin{equation}
%	z^{\textrm{opt}} = \min\left( \argmax_{z}(\text{SC}(\mathcal{Y}, \mathcal{Z}))\right).
%\end{equation}
\begin{align}
	& z^{\textrm{opt}}_{m} = \min_{z_{m} \in \mathcal{C}_{m}} |z_{m}| \nonumber\\
	&\mathcal{C}_{m} \triangleq \argmax_{z_{m}}(\text{SC}(\mathcal{Y}_{m}, \mathcal{Z}_{m}))
	\label{eqn:scmax}
\end{align}
for every dimension $m$. In the sequel, dedicated attack schemes for the $\alpha$-trimmed mean, Talwar and Biweight Tukey are proposed.

\textit{$\alpha$-trimmed mean:} As $\alpha N_{k}$ of largest and smallest weights will be removed, the $(N_{k} - \alpha N_{k} - 1)$-th weight of the ascending sorted weights is chosen. Then the $(N_{k} - \alpha N_{k} - 1)$-th to the $(N_{k} - \alpha N_{k} - P - 1)$-th weights are replaced with the $(N_{k} - \alpha N_{k} - 1)$-th weight. Instead of using the $(N_{k} - \alpha N_{k} - 1)$-th weight, it is also possible to use the $(N_{k} - \alpha N_{k})$-th weight minus some small value. This will have the effect that all malicious weights will be kept and will have a maximal influence on the aggregated weight.

\textit{Talwar and Biweight Tukey:} Both attack schemes will lead to a similar solution, which only differs in a constant $c_0$. By solving
\begin{equation}
	c_0 = \argmax_{x} \psi(x)
\end{equation}
with $\psi(x)$ form Equations~\eqref{eqn:psiTal} and \eqref{eqn:psiTuk}, we obtain for Talwar $c_0 = c$ and for Biweight Tukey $c_0 = \frac{c}{\sqrt{5}}$. As M-estimators subtract a robust location estimate and normalize with a robust scale estimate, we have to account for these shifts by calculating the initial value as
\begin{equation}
	z_{m}^{\textrm{init}} = c_0 \cdot \textrm{mad}(\mathcal{Y}_{m}) + \textrm{median}(\mathcal{Y}_{m})
\end{equation}
where $\textrm{mad}(x)$ denotes the median absolute deviation. Injecting these values into the aggregation step will influence the robust location and scale estimates. Hence, we have to account for this shift by calculating the optimal outlier with  
$\mathcal{Z}_{m}^{\textrm{init}} = \{z_{m}^{\textrm{init}} \cdot \mathds{1}_{|\mathcal{N}^{m}_{k}|}\}$, resulting in
\begin{equation}
	z_{m}^{\textrm{opt}} = c_0 \cdot \textrm{mad}(\mathcal{Y} \cup \mathcal{Z}_{m}^{\textrm{init}}) + \textrm{median}(\mathcal{Y}_{m} \cup \mathcal{Z}_{m}^{\textrm{init}}).
\end{equation}
One might expect this to lead to an iterative calculation, but this is not the case, as the median is primarily influenced by the number of values added to the set and not by the actual values. The resulting values $z_{m}^{\textrm{opt}}$ are shown in Figure~\ref{fig:scMax} as black triangles. Clearly, it would be also possible to use the values mirrored through the origin. Finally, all malicious nodes in the neighborhood of node $k$ will send these maximal influential values and, hence, will cause the largest bias possible in the aggregation. Applying this method over multiple rounds of learning, will lead to a small but steady distortion of the estimates, as we will show in the next section.

%\begin{figure}
%	\centering
%	\resizebox{0.8\columnwidth}{!}{
%		\input{figs/scMax.tex}}
%	\caption{SCs of the presented attack schemes. The black triangles denote the optimal outlier for the respective attack scheme, which maximizes the SC.}
%	\label{fig:scMax}
%    \vspace{-3mm}
%\end{figure}

\begin{figure}
\centering
    \begin{minipage}[t]{.48\columnwidth}
        \centering
    	\resizebox{\columnwidth}{!}{\input{figs/scMax.tex}}
    	\caption{SCs of the presented attack schemes. The black triangles denote the optimal outlier, which maximizes the respective SC.}
    	\label{fig:scMax}
        \vspace{-3mm}
    \end{minipage}%
\hfill
    \begin{minipage}[t]{.48\columnwidth}
	\centering
	\resizebox{\columnwidth}{!}{\input{figs/noOut.tex}}
	\caption{Training loss evolution for all presented aggregation schemes without malicious agents.}
	\label{fig:noOut}
    \vspace{-3mm}
    \end{minipage}
    \vspace{-3mm}
\end{figure}
\vspace{-2mm}

%\begin{figure}[!ht]
%	\removelatexerror
%	\begin{algorithm}[H]
%		\KwIn{$c_0$, $\boldsymbol{\phi}_{l,i}$}
%		\KwOut{$z^{\textrm{max}}$}
%		\For{$m = 1,\dots,M$}
%		{%
%			\[
%				z^{\textrm{init}} = c_0 \cdot \textrm{mad}(\boldsymbol{\phi}_{l,i}) + \textrm{median}(\boldsymbol{\phi}_{l,i})
%			\]
%			\[
%				z^{\textrm{max}} = c_0 \cdot \textrm{mad}([x, z^{\textrm{init}}]) + \textrm{median}([x, z^{\textrm{init}}])
%			\]
%		}
%		\caption{minmax SC estimation for M-estimators}
%		\label{alg:em}
%	\end{algorithm}
%\end{figure}

%% file: figs/scMax.tex
\begin{tikzpicture}
	\begin{axis}[
		width = \figurewidth,
		height = \figureheight,
		xlabel = outlier value,
		ylabel = sensitivity curve,
		grid=both,
		cycle list name=matlabcolor,
		legend pos=north west,
		legend cell align={left},
		]
		\addplot+[thick,matlabyellow,line width=1.5pt,no marks] table[x index = 0,y index=5,col sep=comma] {figs/SC.csv};
		\addplot+[thick,matlabpurple,line width=1.5pt,no marks] table[x index = 0,y index=3,col sep=comma] {figs/SC.csv};
		\addplot+[thick,matlaborange,line width=1.5pt,no marks] table[x index = 0,y index=4,col sep=comma] {figs/SC.csv};
		\addplot+[thick,black,line width=1.5pt, only marks, mark size=4pt] table[x index = 0,y index=1,col sep=comma] {figs/SC_max.csv};
		\legend{Talwar, Tukey, $\alpha$-trimmed mean, $z_{m}^{\textrm{opt}}$};
	\end{axis}
\end{tikzpicture}

%% file: figs/noOut.tex
\begin{tikzpicture}
	\begin{semilogyaxis}[
		width = \figurewidth,
		height = \figureheight,
		xlabel = iteration,
		ylabel = training loss,
		grid,
		xmin=0,
		xmax=300,
		cycle list name=matlabcolor,
		legend pos=north east,
		legend cell align={left},
		]
		\addplot+[thick,line width=1.5pt,no marks] table[skip first n=1, col sep=comma,x index = 0,y index =1] {figs/sim/train_loss_edge_07_mal_0_out_large_value.csv};%
		\addplot+[thick,line width=1.5pt,no marks] table[skip first n=1, col sep=comma,x index = 0,y index =2] {figs/sim/train_loss_edge_07_mal_0_out_large_value.csv};%
		\addplot+[thick,line width=1.5pt,no marks] table[skip first n=1, col sep=comma,x index = 0,y index =3] {figs/sim/train_loss_edge_07_mal_0_out_large_value.csv};%
		\addplot+[thick,line width=1.5pt,no marks] table[skip first n=1, col sep=comma,x index = 0,y index =4] {figs/sim/train_loss_edge_07_mal_0_out_large_value.csv};%
		\addplot+[thick,line width=1.5pt,no marks,legend pos=south east] table[skip first n=1, col sep=comma,x index = 0,y index =5] {figs/sim/train_loss_edge_07_mal_0_out_large_value.csv};%
		\legend{Sample mean,$\alpha$-trimmed mean,Talwar,Tukey,Median}
	\end{semilogyaxis}
\end{tikzpicture}

%% file: sections/simulation.tex
\section{Simulations}
\label{sec:sim}

The simulations are performed on a network with $K=32$ agents arranged in an Erdős–Rényi graph with an edge probability of 70\%. Each agent observes data following a linear model of the form 
\begin{equation}
	\boldsymbol{d}_{k} = \boldsymbol{u}^{\top}_{k} w^{o} + \boldsymbol{v}_{k}
\end{equation}
with the regressors $\boldsymbol{u}_{k} \in \mathbb{R}^{10}$ being independently identical distributed as $\boldsymbol{u}_{k} \sim \mathcal{N}(0, I_{10})$. The noise is distributed as $\boldsymbol{v}_{k} \sim \mathcal{N}(0, \sigma^2_{v})$ with $\sigma^2_{v} = 0.01$. In the simulations the following two assumptions are made, which are commonly found in the literature:
\begin{itemize}
    \item The graph is generated such that for every benign node $k$, the majority of the neighborhood $\mathcal{N}_{k}$ is benign.
    \item Each agent employs a Huber loss function which guarantees $||\nabla J_{k}(w)|| < \infty$.
\end{itemize}
Different numbers of outliers and four different attack schemes were investigated. The deployed attack schemes include: large value (LV), $\alpha$-SCM, Talwar-SCM and Tukey-SCM. Here, LV injects a large constant value, which can be viewed as approaching  the sensitivity curve maximization (SCM) strategy for the sample mean.

\subsection{Efficiency}
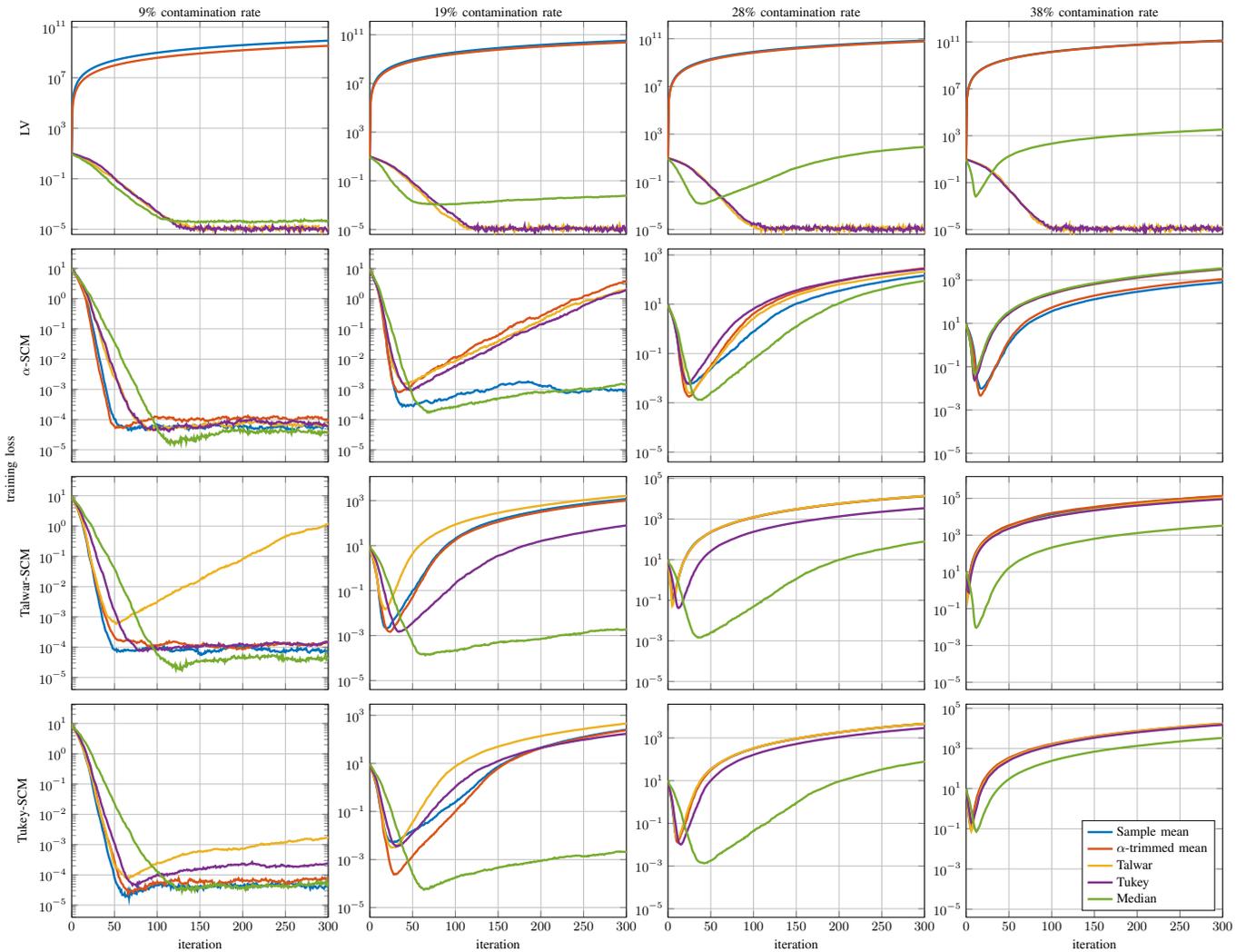
\begin{figure*}[!ht]
	\centering
	\resizebox{2.05\columnwidth}{!}{
		\input{figs/plot_results.tex}}
	\caption{Training loss over iterations. Columns show different amount of outliers. Rows show different attack schemes.}
	\label{fig:results}
    \vspace{-5mm}
\end{figure*}

The asymptotic relative efficiency (ARE) describes the performance loss to the optimal maximum likelihood estimator \cite[p.~22]{Zoubir.2018}. Assuming an underlying normal distribution, the mean has an ARE of $1$ and the median has an ARE of $\frac{2}{\pi} = 0.64$. For the other estimators the ARE can be tuned based on their tuning variables. Here, all aggregators are tuned such that they achieve an efficiency of $0.95$. Hence, we choose $\alpha = 0.0688$  for the $\alpha$-trimmed mean \cite[p.~35]{Mosteller.1977}, $c = 2.7955$ for Talwar \cite{Holland.1977, Menezes.2021} and $c = 4.685$ for biweight Tukey \cite[p.~23]{Zoubir.2018}. This allows for a fair comparison of the different aggregation schemes.

\subsection{Results}

In Figure~\ref{fig:noOut}, the training loss without malicious agents is shown. It can be observed that the sample mean converges fastest and the median slowest, which is consistent with the fact that they exhibit the highest and lowest efficiency respectively. 
%\begin{figure}
%	\centering
%	\resizebox{0.83\columnwidth}{!}{
%		\input{figs/noOut.tex}}
%	\caption{Training loss evolution without malicious agents.}
%	\label{fig:noOut}
%    \vspace{-3mm}
%\end{figure}
The effects of a different amount of outliers and different attack schemes are shown in Figure~\ref{fig:results}. The columns depict different contamination rates, whereas the rows depict different attack schemes. It can be observed that the aggregation scheme which is under attack usually has the worst performance or a significantly worse performance, compared to other attack schemes. For example, in the second column under the LV attack the worst performance is shown by the sample mean, under $\alpha$-SCM the worst performance is shown by the $\alpha$-trimmed mean, under Talwar-SCM the worst performance is shown by the Talwar aggregator and under Tukey-SCM the worst performance is also shown by the Talawr aggregator, but closely followed by the Tukey aggregator, which shows a worse performance compared to the other attack schemes.
Even under moderate contamination rates the aggregation schemes fail, if a targeted SCM is applied. We conclude that carefully targeted attacks can be effective even against aggregation schemes which are robust to simpler attack models.

%% file: figs/plot_results.tex
\begin{tikzpicture}
	\begin{groupplot}[
			scale=0.9,
			group style={group name=my plots,
				group size= 4 by 4,
				xlabels at=edge bottom,
				xticklabels at=edge bottom,
				%ylabels at=edge left,
				%yticklabels at=edge left,
				vertical sep=10pt},
			cycle list name=matlabcolor,
			grid,
			xlabel=iteration,
			xmin = 0,
			xmax = 300,
            ymin= 0.4e-5,
			legend style={
				legend pos= south east,
				legend cell align={left},
			},
			]
			\def\myPlots{}
			\pgfplotsforeachungrouped \myA in {{large_value}, {max_alpha_mean}, {max_talwar_mean}, {max_tukey}}{
				\pgfplotsforeachungrouped \myB in {3,6,9,12}{
					\eappto\myPlots{%
						\noexpand\nextgroupplot[ymode=log]
						\noexpand\addplot+[thick,line width=1.5pt,no marks] table[skip first n=1, col sep=comma,x index = 0,y index =1] {figs/sim/train_loss_edge_07_mal_\myB_out_\myA.csv};%
						\noexpand\addplot+[thick,line width=1.5pt,no marks] table[skip first n=1, col sep=comma,x index = 0,y index =2] {figs/sim/train_loss_edge_07_mal_\myB_out_\myA.csv};%
						\noexpand\addplot+[thick,line width=1.5pt,no marks] table[skip first n=1, col sep=comma,x index = 0,y index =3] {figs/sim/train_loss_edge_07_mal_\myB_out_\myA.csv};%
						\noexpand\addplot+[thick,line width=1.5pt,no marks] table[skip first n=1, col sep=comma,x index = 0,y index =4] {figs/sim/train_loss_edge_07_mal_\myB_out_\myA.csv};%
						\noexpand\addplot+[thick,line width=1.5pt,no marks,legend pos=south east] table[skip first n=1, col sep=comma,x index = 0,y index =5] {figs/sim/train_loss_edge_07_mal_\myB_out_\myA.csv};%
					}
				}
			}
			\myPlots
			\legend{Sample mean,$\alpha$-trimmed mean,Talwar,Tukey,Median}
	\end{groupplot}
	\node[anchor=south] at ($(my plots c1r1.north)$){9\% contamination rate};
	\node[anchor=south] at ($(my plots c2r1.north)$){19\% contamination rate};
	\node[anchor=south] at ($(my plots c3r1.north)$){28\% contamination rate};
	\node[anchor=south] at ($(my plots c4r1.north)$){38\% contamination rate};
	\node[anchor=south, rotate=90, yshift=9mm] at ($(my plots c1r1.west)$){LV};
	\node[anchor=south, rotate=90, yshift=9mm] at ($(my plots c1r2.west)$){$\alpha$-SCM};
	\node[anchor=south, rotate=90, yshift=9mm] at ($(my plots c1r3.west)$){Talwar-SCM};
	\node[anchor=south, rotate=90, yshift=9mm] at ($(my plots c1r4.west)$){Tukey-SCM};
	\node[anchor=south, rotate=90, yshift=12mm] at ($(my plots c1r2.west)!0.5!(my plots c1r3.west)$){training loss};
\end{tikzpicture}

%% file: sections/conclusion.tex
\section{Conclusion and Outlook}
\label{sec:con}
We presented a novel attack on decentralized learning algorithms by injecting an outlier which maximizes the sensitivity curve of the used aggregation scheme. The simulations show that all considered aggregation schemes diverge. These results suggest that many robust aggregation schemes, despite their short-term robustness, can be driven to divergence through small but targeted perturbations.

Efficient defenses against the presented SC maximization attack might be a stronger reliance on each nodes own estimated weights, e.g. self-centered clipping \cite{He.2022}. Another approach could be the obfuscation of the used aggregation scheme, e.g. by randomly choosing a scheme out of a predefined set \cite{AliRamezaniKebrya.2022}. This could be combined with randomly adapting the tuning variables, which would make it more difficult for malicious agents to design optimal attack patterns.